%% file: main.tex
\documentclass[10pt,twocolumn,letterpaper]{article}

\usepackage{wacv}              


\definecolor{wacvblue}{rgb}{0.21,0.49,0.74}
\usepackage[pagebackref,breaklinks,colorlinks,allcolors=wacvblue]{hyperref}
\usepackage{multirow}

\usepackage{tcolorbox}
\tcbuselibrary{breakable}
\usepackage{algorithm}
\usepackage{algpseudocode}

\def\wacvPaperID{*} 
\def\confName{WACV}
\def\confYear{2027}

\title{Generative Spatiotemporal Data Augmentation via Video Diffusion}

\author{
Jinfan Zhou$^{*1,2}$ \quad
Lixin Luo$^{*2}$ \quad
Sungmin Eum$^{3}$ \quad
Heesung Kwon$^{3}$ \quad
Jeong Joon Park$^{2}$ \\
\small $^{1}$University of Chicago \quad
$^{2}$University of Michigan, Ann Arbor \quad
$^{3}$DEVCOM Army Research Laboratory \\
{\tt\small zjf@uchicago.edu} \quad
{\tt\small \{lixinluo,jjparkcv\}@umich.edu} \quad
{\tt\small \{sungmin.eum,heesung.kwon\}.civ@army.mil} \\
}

\begin{document}

\twocolumn[{%
\renewcommand\twocolumn[1][]{#1}%
\maketitle
\vspace{-10pt}
\includegraphics[width=\linewidth]{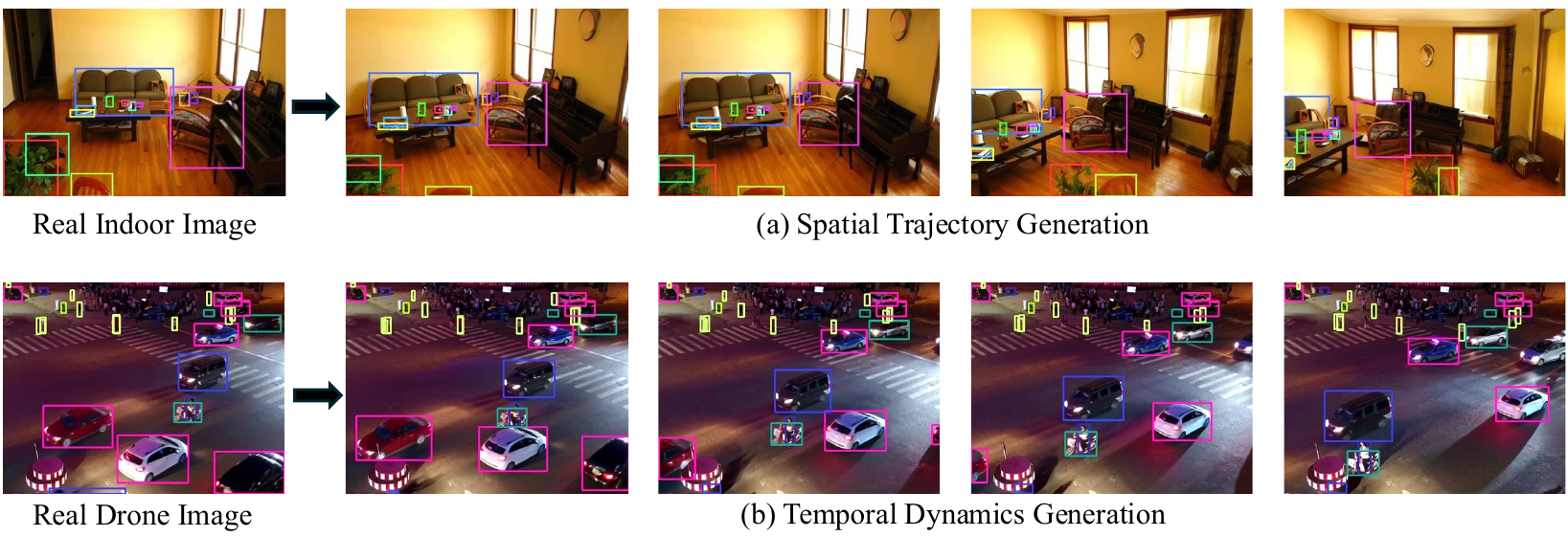}
\captionof{figure}{
    \textbf{Spatiotemporal augmentation from a single image}. Given a single real image (left), our method generates diverse spatial viewpoints (top row) and plausible temporal dynamics (bottom row) using video diffusion models. Bounding boxes are automatically propagated to all synthetic frames, enabling training with rich geometric and motion variation without additional manual annotation cost.
    \vspace{0.5cm}
}
\label{fig:teaser}
}]


\input{sec/0_abstract}    
\input{sec/1_intro}
\input{sec/2_related_works}
\input{sec/3_methods}
\input{sec/4_experiments}
\input{sec/5_conclusion}
\paragraph*{Acknowledgements} This research was supported by the Army Research Office (ARO) under Grant No. W911NF25-1-0047.
{
    \small
    \bibliographystyle{ieeenat_fullname}
    \bibliography{main}
}

\appendix
\input{sec/X_suppl}

\end{document}

%% file: sec/0_abstract.tex

{\centering\large\bfseries Abstract\par}
{\itshape
We explore spatiotemporal data augmentation using video foundation models to diversify both camera viewpoints and scene dynamics. Unlike existing approaches based on simple geometric transforms or appearance perturbations, our method leverages off-the-shelf video diffusion models to generate realistic 3D spatial and temporal variations from a given image dataset. Incorporating these synthesized video clips as supplemental training data yields consistent performance gains in low-data settings, such as UAV-captured imagery where annotations are scarce. Beyond empirical improvements, we provide practical guidelines for choosing an appropriate spatiotemporal generative setup and transferring annotations to synthetic frames. Experiments on COCO subsets and UAV-captured datasets show that, when applied judiciously, spatiotemporal augmentation broadens the data distribution along axes underrepresented by traditional and prior generative methods, offering an effective lever for improving model performance in data-scarce regimes.
}


%% file: sec/1_intro.tex
\section{Introduction}
\label{sec:intro}

Deep learning has driven remarkable progress in core vision tasks such as image classification~\cite{krizhevsky2012imagenet, he2016deep, dosovitskiy2020image}, object detection~\cite{girshick2015fast, redmon2016you, carion2020end, li2022exploring}, and semantic segmentation~\cite{ronneberger2015u, chen2017rethinking, kirillov2023segment}.
However, this progress relies on large, diverse, and densely annotated datasets~\cite{deng2009imagenet, lin2014microsoft, singh2024benchmarking}.
Collecting new viewpoints, camera motions, and object dynamics, especially in domains like drones, robotics, or long-tail scenes, often requires specialized hardware or repeated deployments, and manual annotation remains labor-intensive.

These challenges have motivated the use of generative models~\cite{goodfellow2014generative, oord2016pixel, ho2020denoising} to synthesize training data.
Diffusion models now produce photorealistic images~\cite{rombach2022high}, 3D assets~\cite{poole2022dreamfusion}, and high-quality videos~\cite{blattmann2023stable}.
Yet despite this fidelity, prior work largely focuses on appearance-level augmentation, leaving the question of how to leverage generative models to meaningfully expand a dataset’s spatiotemporal diversity.

In this work, we show that modern video diffusion models encode strong, implicit 3D and dynamic priors~\cite{yu2024viewcrafter, sun2024dimensionx, zhou2025stable, hacohen2024ltx} that can be directly exploited for perception tasks.
From a single input image, these models generate novel viewpoints and plausible temporal variations, yielding data with richer geometric and temporal coverage than the original dataset.
Because the frames are temporally coherent, visual foundation models~\cite{kirillov2023segment, ravi2024sam} can propagate annotations reliably across the generated sequence.

Our key idea is to demonstrate that an off-the-shelf video diffusion model  acts as a powerful spatiotemporal data augmentation engine for perception tasks, especially with scarce data.
We provide practical guidelines, backed by extensive experiments, for how to use this augmentation effectively.
This approach yields significant gains in data-scarce detection regimes: +3.8 mAP on a COCO \cite{lin2014microsoft} sub-dataset, +2.8 mAP on VisDrone \cite{cao2021visdrone}, and +5.9 mAP on Semantic Drone \cite{Charles2013}. Moreover, our experiments on instance segmentation and self-supervised training (DINO \cite{caron2021dino}) further show significant benefits across diverse perception tasks.
To enable scalable deployment, we introduce an automated annotation pipeline that transfers bounding boxes across generated views, handles disocclusions, and offers tools for balancing coverage and fidelity.
Our contributions are:

    1. A first demonstration that an off-the-shelf video diffusion model can generate synthetic data that measurably improves object detection performance.

    2. A fully automated annotation-transfer pipeline that leverages temporal consistency to produce accurate annotations for all generated frames.

    3. Extensive experiments offering guidelines on choosing video diffusion types, number of frames, camera trajectories, disocclusion handling, and other factors that influence the effectiveness of spatiotemporal augmentation.

\begin{figure*}[htbp]
    \centering
    \includegraphics[width=1.0\textwidth]{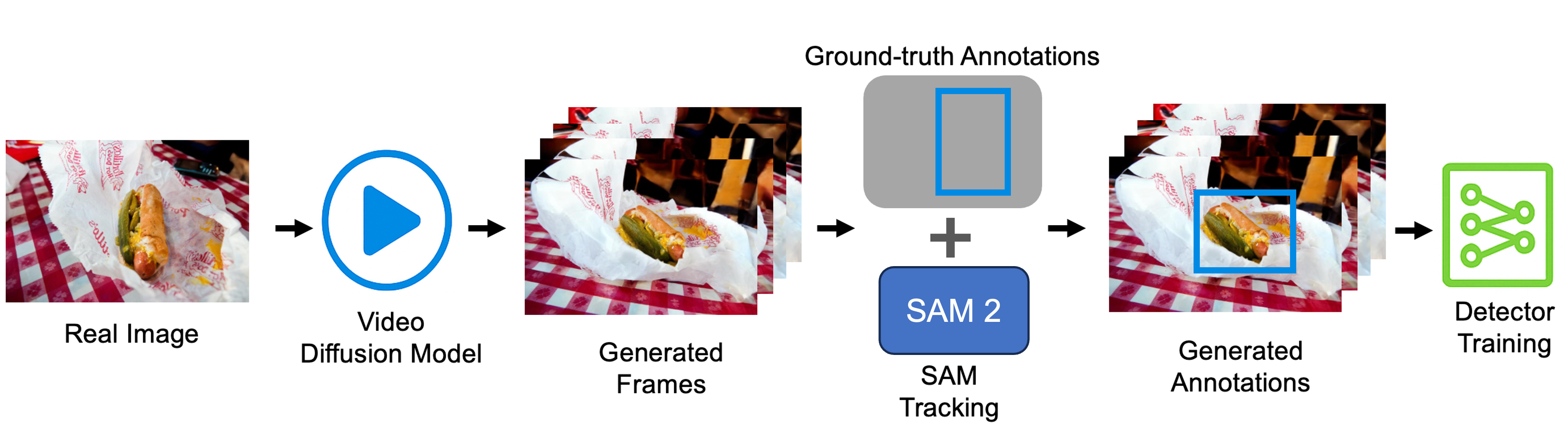}
    
    \caption{We use video diffusion models to expand spatially and temporally to the current scene. Thanks to the consecutive frames of the generated video, SAM 2 video segmentation model can be used to robustly, accurately and automatically generate segmentation masks and the bounding boxes for the generated frames.}
    
    \label{fig:pipe1}
\end{figure*}

%% file: sec/2_related_works.tex
\section{Related Works}
\label{related}

\subsection{3D-aware Video Diffusion Models}

The remarkable success of diffusion models in image synthesis~\cite{ho2020denoising, song2020denoising, song2020score, karras2022elucidating} has been rapidly extended to the temporal domain of video generation. Foundational text-to-video models such as Video Diffusion Models~\cite{ho2022video} and Make-A-Video~\cite{singer2022make}, along with more recent large-scale systems~\cite{yang2024cogvideox, wan2025wan, hacohen2024ltx}, have demonstrated the ability to generate high-fidelity and temporally coherent video clips from textual descriptions. However, precise control over the camera viewpoint remains a key challenge. To address this, one line of work~\cite{yu2024viewcrafter} leverages explicit geometric condition to guide generation. Another line of methods~\cite{zhou2025stable, sun2024dimensionx} learns to interpret camera parameters directly, bypassing the need for an intermediate 3D representation. The development of these controllable models provides powerful, off-the-shelf tools for generating visually consistent, multi-view data~\cite{wu2025cat4d, bahmani2025ac3d}. In this paper, we investigate how the rich, 3D-aware priors embedded in these models can be directly harnessed to improve the performance of downstream perception tasks.

\subsection{Generative Data augmentation}
Before the advent of generative models, sophisticated data augmentation methods were proposed to synthesize new images~\cite{zhang2017mixup, yun2019cutmix, zhong2020random}. With the recent blossoming of generative models—renowned for producing high-quality, high-fidelity, and diverse outputs—a growing number of methods now leverage them for data augmentation~\cite{fang2024data, he2022synthetic, trabucco2023effective, vu2025multi, zhu2024odgen}. For instance, He et al.~\cite{he2022synthetic} extensively study how synthetic images from text-to-image models can be used for image recognition. Trabucco et al.~\cite{trabucco2023effective} edit image semantics with an off-the-shelf diffusion model to generalize from few-shot examples, while Fang et al.~\cite{fang2024data} use visual priors to guide generation and employ category-calibrated CLIP scores for post-filtering.

These works primarily focus on diversifying the visual appearance of existing images. In contrast, our method leverages the temporal and spatial priors learned by video diffusion models to expand the scene spatially by generating new viewpoints and temporally by converting the static image into a dynamic video. Furthermore, to obtain reliable annotations for training an object detector, previous approaches either perform location-invariant augmentations~\cite{trabucco2023effective} or generate objects conditioned on given bounding boxes~\cite{zhu2024odgen, vu2025multi}. Our method, however, generates consecutive video frames and employs a foundation model~\cite{ravi2024sam} for automatic annotation, thereby allowing for greater flexibility in the location of the target objects.

\subsection{UAV-based Object Detection}
Object detection from Unmanned Aerial Vehicles (UAVs) introduces unique challenges not present in ground-based detection~\cite{zhu2021detection, du2018unmanned, wu2019delving, hsieh2017drone, benchmark2016benchmark, barekatain2017okutama}. Key research challenges in this domain stem from the nature of aerial imagery. Objects frequently appear at small scales, in dense clusters, or are sparsely distributed, requiring specialized architectures that fuse multi-resolution features~\cite{liu2020hrdnet, liu2020small} or employ cluster-based proposal mechanisms~\cite{yang2019clustered}. Furthermore, constant variations in UAV altitude and viewing angles create significant appearance shifts.

For practical deployment, computational efficiency is paramount for on-board processing. This has driven a shift from slower two-stage methods towards efficient one-stage detectors, represented by YOLO ~\cite{redmon2016you, yolo11_ultralytics}. Recently, transfomer-based DETR models have also emerged as a powerful family of object detectors~\cite{carion2020end, lv2023detrs}. These models offer a favorable balance of speed and accuracy, making them a common choice for UAV-based detection systems~\cite{liu2020uav}.

%% file: sec/3_methods.tex
\section{Methods}
\label{methods}

In this section, we first formalize our problem and introduce our generative spatiotemporal augmentation pipeline. We then analyze how the proposed augmentation expands data coverage, followed by implementation details on annotation transfer and disocclusion handling.

\subsection{Generative Spatiotemporal Augmentation}

We aim to enrich training data by generating new views and dynamics from existing images using a pre-trained image-to-video diffusion model $\mathcal{G}$. Suppose a dataset consisting of $N$ images $\mathcal{D}_{\text{orig}} = \{I^{(i)}_0\}_{i=1}^N$, for each original image $I^{(i)}_0$ in the dataset, we generate a video sequence $\{I^{(i)}_t\}_{t=1}^T$ that depicts alternative viewpoints or motions of the same scene, where $T$ is the number of generated frames. By conducting this augmentation to all $N$ images, we obtain the synthetic dataset $\mathcal{D}_{\text{new}}=\{\{I^{(i)}_t\}_{t=1}^T\}_{i=1}^N$, and the final augmented dataset  $\mathcal{D}_{\text{aug}}=\mathcal{D}_{\text{orig}} \cup \mathcal{D}_{\text{new}}$. 


\paragraph{3D Novel-View Augmentation.}

To simulate multi-view geometry, we synthesize novel camera viewpoints of a (quasi-)static scene. Let the camera trajectory be a sequence of poses
$C=\{c_t\}_{t=1}^T$, where each $c_t \in SE(3)$ specifies the rotation and translation relative to the initial view. The video model $\mathcal{G}$ generates frames as
\[
\{I_t\}_{t=1}^T = \mathcal{G}(I_0, C).
\]
We then select a subset of frames indexed by $S \subseteq \{1,\ldots,T\}$, $|S|=K$, to form
\[
\mathcal{D}_{\text{new}} = \{ I_t \mid t \in S \}.
\]
This turns a single 2D image into geometrically diverse views guided by the 3D priors of $\mathcal{G}$. 

\paragraph{Temporal-Dynamics Augmentation.}

To further diversify temporal patterns, we prompt $\mathcal{G}$ to synthesize dynamic scene evolution:
\[
\{I_t\}_{t=1}^T = \mathcal{G}(I_0; \tau),
\]
where $\tau$ is a temporal control signal (e.g., motion strength or duration). We sub-select frames $S \subseteq \{1,\ldots,T\}$ to augment $\mathcal{D}_{\text{orig}}$, yielding $\mathcal{D}_{\text{aug}}$. We evaluate different video diffusion backbones, spatial cropping schemes, and frame-selection strategies to maximize augmentation utility while avoiding degenerate 
samples.

\subsection{Coverage Expansion Analysis}
\label{sec:coverage}
Our central hypothesis is that generative spatiotemporal augmentation expands the effective coverage of a limited training set over the unseen target distribution. Unlike classical flips or crops, which perturb images locally, our method explores underrepresented axes of variation—multi-view geometry and temporal dynamics—thus enriching the data manifold along directions that standard augmentations neglect.

\paragraph{Quantifying Coverage.}
We adopt the precision--recall framework for generative models~\cite{kynkaanniemi2019improved} to measure how an augmented training set \( D \) spans a held-out validation set \( U \).
Given a feature encoder \( \phi(\cdot) \) and distance metric \( d(\cdot,\cdot) \), recall is defined as
\begin{equation}
f(I, D) =
\begin{cases}
1, & \text{if } \exists I' \in D \text{ s.t. } d(\phi(I), \phi(I')) \le r(I', D),\\
0, & \text{otherwise,}
\end{cases}
\end{equation}
where \( r(I', D) \) is the distance to the \( k \)-th nearest neighbor of \( I' \) within \( D \).
The overall recall is
\begin{equation}
\mathrm{Recall}(U \!\rightarrow\! D) = 
\frac{1}{|U|} \sum_{I \in U} f(I, D),
\end{equation}
representing the fraction of validation samples whose features are covered by the training set.
Following~\cite{kynkaanniemi2019improved}, we use a pretrained VGG-16~\cite{simonyan2014very} with \( k=3 \) for all measurements.


\paragraph{Coverage Growth and Diminishing Returns.}
As the number of synthetic frames increases, both image- and object-level recall rise sharply, confirming that the generated samples fill underrepresented regions of the data manifold.
The recall curve typically saturates near 0.93 after approximately 30 frames, indicating that our augmentation efficiently reaches the dominant modes of variation without excessive sampling.

\begin{figure}[t]
    \centering
    \includegraphics[width=\linewidth]{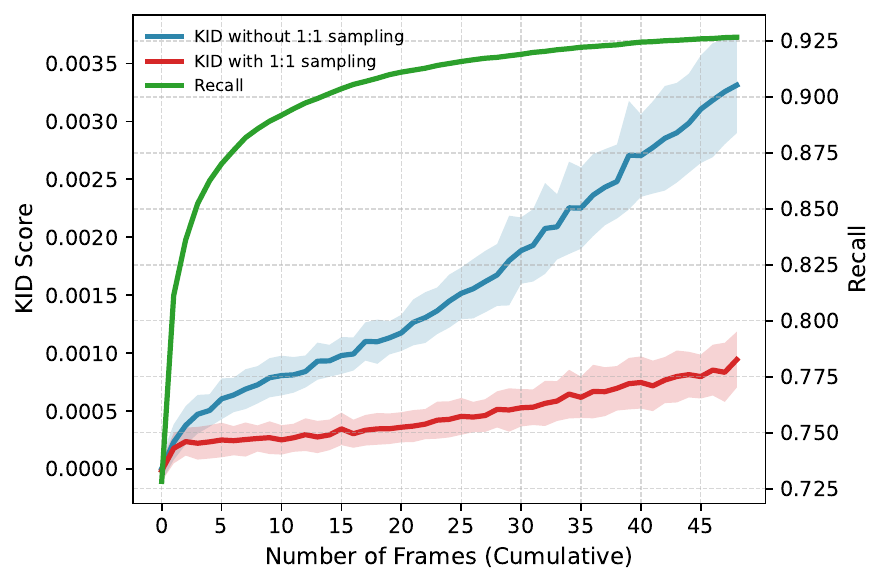}
    \caption{
    \textbf{Object recall and KID vs.\ number of synthetic frames.}
    Recall increases sharply then saturates after $\sim$30 frames, showing diminishing returns in coverage. 
    KID increases with frame count, but 1:1 sampling (red) stabilizes realism compared to unrestricted sampling (blue).
    }
    \label{fig:recall_vs_frames}
\end{figure}

\begin{table}[t]
\centering
\caption{\textbf{Comparison of image and object recall} under different augmentations. Results are computed using 8 generated images per original image with a 1:1 mixing ratio for all augmentation methods. }
\resizebox{\columnwidth}{!}{
\begin{tabular}{ccccc}
\toprule
Augmentation & Raw  & Instagen & ControlAug & Ours \\
\midrule
Image Recall  & 88.60  & 97.14 & 98.62 & \textbf{98.98} \\
Object Recall & 72.83  & 78.42 & 86.46 & \textbf{88.56} \\
KID$\downarrow$ & $5 \times 10^{-6}$  & $5.5 \times 10^{-3}$ & $8.7 \times 10^{-4}$ & $\mathbf{6.5 \times 10^{-4}}$ \\
\bottomrule
\end{tabular}
}
\label{tab:recall_comparison}
\end{table}





\paragraph{Complementarity with Existing Augmentations.}
While our method alone outperforms appearance-based augmentations such as Instagen~\cite{feng2024instagen} and ControlAug~\cite{fang2024data} in terms of coverage (Tab.~\ref{tab:recall_comparison}), further performance gains can be obtained by combining spatiotemporal augmentations with existing methods (see supplementary).
This suggests that our approach expands the training distribution along axes that prior methods underexplore.
Moreover, in compute- or data-limited settings, the coverage efficiency of our augmentation offers a practical way to achieve diversity without retraining or extensive manual tuning.

\paragraph{Coverage--Fidelity Trade-off.}
Because video diffusion models produce temporally correlated frames, precision metrics are less informative.
We therefore use the Kernel Inception Distance (KID)~\cite{binkowski2018demystifying} to quantify the distribution shift in the training data. 
As shown in Fig.~\ref{fig:recall_vs_frames}, recall improves steadily as KID increases mildly, revealing a coverage--fidelity trade-off.
A 1:1 mixing ratio between real and generated images achieves the best balance, expanding coverage substantially while maintaining overall realism. 
Tab.~\ref{tab:recall_comparison} further confirms this Recall-KID tradeoff across different augmentation methods.

\subsection{Annotation Transfer via Video Segmentation}
To use the synthetic images $S$ for training an object detector or instance segmentation model, we require annotations for each frame. Manually labeling these frames is prohibitively expensive, so we adopt an automated annotation pipeline.

Let $B_0^{j}$ denote the ground-truth bounding box and $M_0^{j}$ denote the ground-truth instance mask for object $j$ in the original image $I_0$. We employ a state-of-the-art video instance segmentation model $\mathcal{M}_{\text{seg}}$ (e.g., SAM 2~\cite{ravi2024sam} tracking), which tracks and segments objects throughout a video given an initial prompt.

For object detection, we prompt the model with the initial bounding box $B_0^{j}$. Given the generated video $V=\{I_t\}_{t=1}^T$, the model produces a sequence of binary masks,
\[
\{M_t^{j}\}_{t=1}^T \;=\; \mathcal{M}_{\text{seg}}(V, B_0^{j}),
\]
where $M_t^{j}$ is the segmentation mask for object $j$ in frame $I_t$. From each mask, we compute the tightest enclosing axis-aligned bounding box,
\[
B_t^{j} \;=\; \text{BoundingBox}\!\left(M_t^{j}\right),
\]
which serves as the annotation for the corresponding synthetic image.

For instance segmentation, we instead prompt the model with the ground-truth mask $M_0^{j}$. The tracker propagates the mask across the generated video,
\[
\{M_t^{j}\}_{t=1}^T \;=\; \mathcal{M}_{\text{seg}}(V, M_0^{j}),
\]
and the propagated mask $M_t^{j}$ is directly used as the instance segmentation label for frame $I_t$.

This process is applied to all objects in $I_0$, yielding a complete set of annotations for the entire generated sequence. Fig.~\ref{fig:teaser} and \ref{fig:pipe1} show examples and a description of this process.

\begin{figure*}[t]
    \centering
    \includegraphics[width=\linewidth]{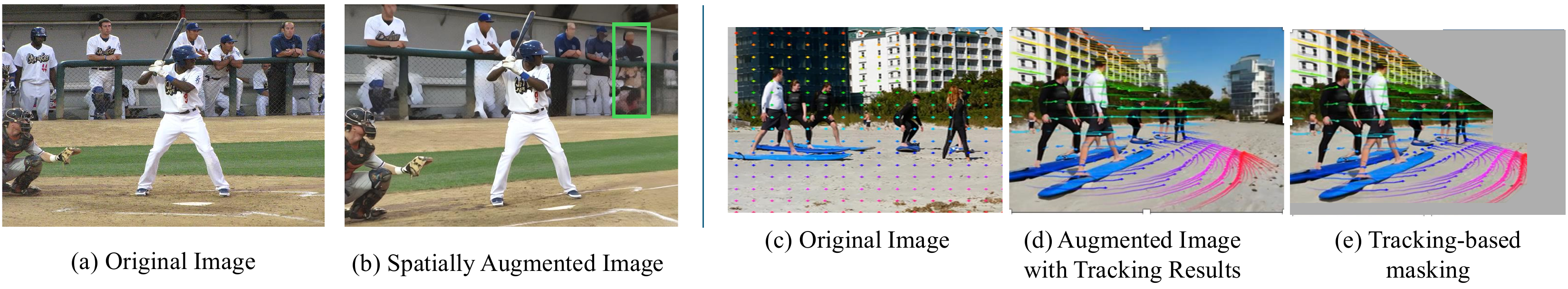}
    \caption{
    \textbf{Disocclusion handling strategies.}
        (a)~Original image. (b)~Spatially augmented image showing newly visible regions (pseudo-label example).
    (c--e)~Illustration of tracking-based masking:
    (c)~original frame with grid points,
    (d)~augmented frame with tracked points, and
    (e)~masked result excluding disoccluded areas.
    These visualizations highlight that disoccluded regions can be detected and masked, though the overall performance benefit is marginal.
    }
    \label{fig:disocclusion}
\end{figure*}

\paragraph{Disocclusion handling.} A practical challenge in our spatiotemporal augmentation is \emph{disocclusion}--- as the virtual camera moves, regions that were previously hidden in the original image become newly visible (see Fig.~\ref{fig:disocclusion}). To address this problem, we explored two strategies: (i) \emph{tracking-based masking}, which densely tracks pixels from the first frame and masks out regions that are not tracked (i.e., newly generated areas), and (ii) \emph{pseudo-labeling}, where a detector trained on the original dataset predicts labels for the newly generated regions. However, our experiments show that both strategies provide negligible benefits, and we therefore omit them from the final pipeline. See supplementary Sec.~3.2 for quantitative results on disocclusion handling.

\subsection{Choosing an Effective Configuration}
\label{sec:config}

Our spatiotemporal augmentation framework exposes several controllable factors for the final performance.
The most important include: 
(1) the augmentation type (3D novel views or temporal dynamics), 
(2) the camera trajectory and number of generated frames, 
(3) frame filtering strategies.

Guided by the Recall–KID analysis (Sec.~\ref{sec:coverage}), we find that effective configurations expand coverage while preserving perceptual fidelity.
In practice, a 1:1 ratio between real and generated images, modest camera motion, and CLIP-filtered frames provide a good balance between realism and diversity.
Comprehensive ablations in Sec.~\ref{sec:experiments} confirm and provides further details of the controllable factors.

\subsection{Generative Augmentation for Self-Supervised Learning}
\label{sec:ssl}

Our spatiotemporal framework naturally extends to augmentation-based self-supervised learning (SSL) methods like MoCo~\cite{he2020moco} and DINO~\cite{caron2021dino}. Standard SSL enforces consistency between randomly transformed views, $x_1 = \mathcal{A}_1(x), x_2 = \mathcal{A}_2(x)$, where $x$ is the original image and $\mathcal{A}_1, \mathcal{A}_2 \sim \mathcal{A}$ denote random transformations sampled from an augmentation distribution. 
Instead, we sample two frames from a generated video sequence conditioned on the original image $I_0 = x$. Let $V = \{ I_t \}_{t=0}^{T} \sim p(V \mid I_0)$ be the synthesized video. 
Two views are constructed by sampling two frames and then applying standard image augmentations:
\[
x_1 = \mathcal{A}_1(I_a), \quad x_2 = \mathcal{A}_2(I_b), 
\]
where \(
a,b \sim \mathcal{U}\{0,\dots,T\}, \quad \mathcal{A}_1,\mathcal{A}_2 \sim \mathcal{A}.
\)
By sampling from the same video, we introduce richer geometric and temporal diversity than traditional image-level augmentation. This preserves semantic consistency while effectively transferring the generative model's multi-view priors into representation learning.



%% file: sec/4_experiments.tex
\section{Experiments}
\label{sec:experiments}

\begin{figure*}[t]
    \centering
    \includegraphics[width=\linewidth]{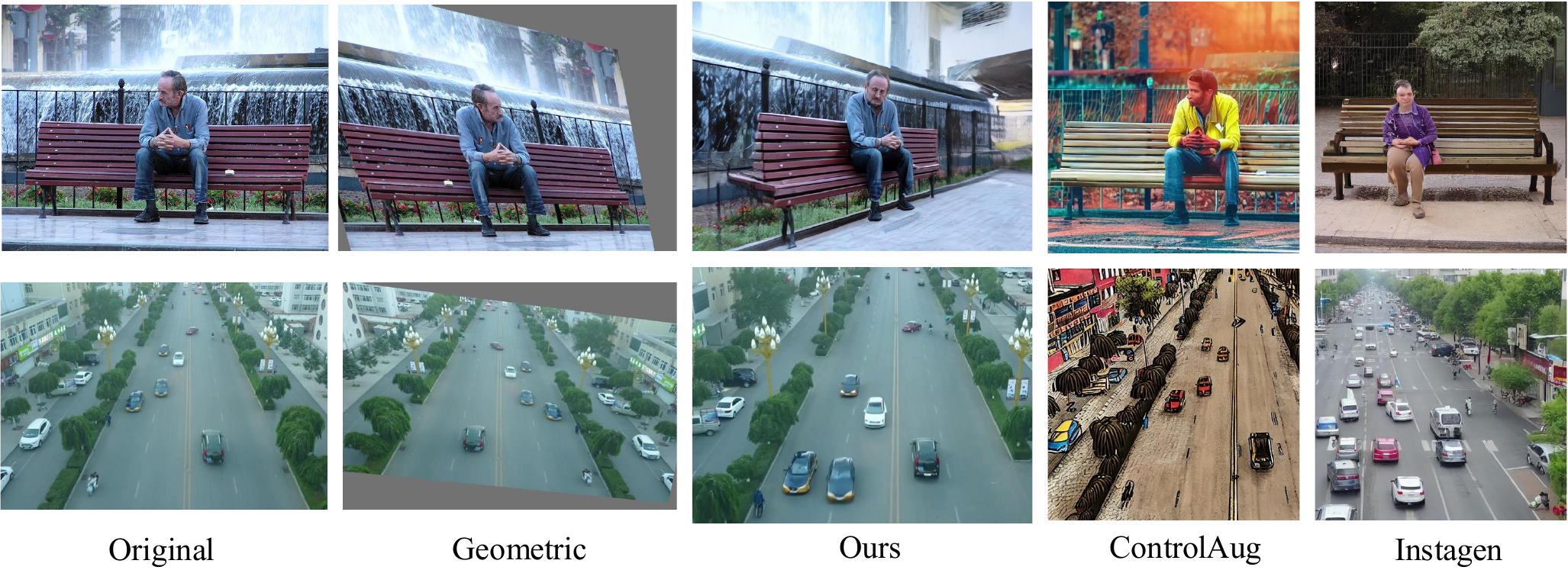}
    \caption{\textbf{Qualitative comparison of our spatiotemporal augmentation} on COCO and VisDrone against baselines.
Our method produces video frames conditioned on the input image, yielding coherent and expressive spatial(top row) and temporal(bottom row) variation compared to the conventional geometric augmentation. ControlAug \cite{fang2024data} synthesizes images conditioned on the HED boundary map of the original image, preserving spatial structure but altering appearance. InstaGen \cite{feng2024instagen} samples new images using a finetuned diffusion model guided by text prompts (e.g., “a photograph of person and bench,” or “a drone photograph containing car, pedestrian, and bicycle”).}
    \label{fig:qualatative-coco}
\end{figure*}

We evaluate spatiotemporal augmentation across several dimensions and evaluate using three datasets including COCO~\cite{lin2014microsoft}, VisDrone~\cite{cao2021visdrone} and Semantic Drone~\cite{Charles2013}. We first compare against traditional baseline and other generative augmentation methods (see Fig.~\ref{fig:qualatative-coco}).
We then analyze how different augmentation types (spatial trajectories vs.\ temporal dynamics) interact with dataset characteristics and the scaling of benefits with dataset size.


\subsection{Spatiotemporal Augmentation Improves Performance}

A central question is whether the additional coverage provided by spatiotemporal augmentation, as shown in Sec.~\ref{sec:coverage}, translates into measurable detection gains. 
We therefore evaluate on COCO5k (a subset of COCO containing 5k samples) and VisDrone (6,471 training images). For COCO, we use DimensionX \cite{sun2024dimensionx} as our video diffusion model; For VisDrone, we use LTX-Video-13B~\cite{hacohen2024ltx} with randomly cropped regions of the original images as conditioning inputs. Because aerial images contain many small objects, cropping improves generation quality. From each image, we extract 5 random crops and generate a short video clip for each. We then sample 8 frames per clip, producing 40 augmented images per original image. As shown in Tab.~\ref{tab:aug_comparison}, our generative augmentation provides significant improvements against all baselines augmentation methods.

We compare our spatiotemporal augmentation against both traditional geometric augmentation and recent generative augmentation baselines InstaGen~\cite{feng2024instagen} and ControlAug~\cite{fang2024data}. For fairness, we fix the number of generated samples across all methods, using 8 augmented images per original image. For InstaGen, annotations for synthetic images are produced using a detector pretrained on the corresponding raw dataset. For ControlAug, we generate 27 images per original sample and keep the top eight according to the CLIP-based filtering protocol proposed in the original work, corresponding to a filtering ratio of 0.3.

Table~\ref{tab:aug_comparison} reports mAP scores for object detection. 
Our generative spatiotemporal data augmentation outperforms all baselines on all datasets.
On VisDrone, a UAV-captured dataset characterized by viewpoint variation and limited data, the improvements are significant.
When fixing the number of augmented frames to 8x, our method also shows better performance than both baselines. 

Also, as a complementary dimension to appearance-based augmentation methods, combining spatiotemporal and appearance-based methods leads to further performance improvements when the amount of added data is fixed (see Tab.~\ref{tab:combined-aug}). These results suggest that our approach expands the training distribution along dimensions that prior appearance-based methods underexplore. By introducing spatiotemporal variations, our method complements existing augmentations that primarily modify appearance, leading to improved data diversity and overall detection performance.



\paragraph{Results on Semantic Drone.}
To further evaluate the effectiveness of our data augmentation approach on smaller datasets, we conduct experiments on the Semantic Drone dataset \cite{Charles2013} for UAV-view person detection. This dataset consists of 400 high-resolution images captured from a nadir view.
We train a YOLO11x model on a subset containing 200 training samples, and use the remaining 200 images for validation.
To construct our augmented dataset, we randomly crop 10 sub-images from each original image to generate 10 short video clips with LTX-Video-13B \cite{hacohen2024ltx}. We uniformly sample 8 frames per video clip, yielding 80 augmented images per original image. 
As shown in Table~\ref{tab:aug_comparison}, our method  outperforms the baseline augmentation methods with massive improvements. 

\paragraph{Results on Instance Segmentation.} Table~\ref{tab:seg} reports Mask mAP for instance segmentation. Our generative spatiotemporal augmentation consistently improves performance across all dataset sizes. The gains are most pronounced in the low-data regime, improving Mask mAP by +3.5 with 5k images. The improvements remain consistent as the dataset grows, still achieving gains of +2.0 Mask mAP over traditional augmentation with 20k images.





\begin{table}[t] 

\caption{\textbf{Augmentation Effectiveness Comparison.}
We compare our method against traditional augmentations, InstaGen~\cite{feng2024instagen}, and ControlAug~\cite{fang2024data} on COCO5k, VisDrone, and Semantic Drone, using the same number of generated frames (8x the original dataset size on COCO5k and VisDrone, 80x on Semantic Drone). Overall, our method outperforms the baselines.}

\label{tab:aug_comparison}
\begin{center}
\resizebox{\columnwidth}{!}{
\begin{tabular}{l|l|c|c}
\hline
Dataset & Method & mAP@50 & mAP@50:95 \\
\hline
\multirow{4}{*}{COCO5k} & Trad & 36.3 & 24.8 \\
& Trad + ControlAug (8x) & 40.1 ($+3.8$)  & 27.6 ($+2.8$) \\
& Trad + InstaGen (8x) & 40.3 ($+4.0$)  & 28.1 ($+3.3$) \\
& Trad + Ours (8x) & \textbf{41.1} ($+\textbf{4.8}$) & \textbf{28.6} ($+\textbf{3.8}$) \\
\hline
\multirow{4}{*}{VisDrone} & Trad  & 45.9 & 28.5 \\
& Trad + ControlAug (8x) & 48.5 ($+2.6$) & 30.5 ($+2.0$) \\
& Trad + InstaGen (8x) & 47.7 ($+1.8$) & 29.8 ($+1.3$) \\
& Trad + Ours (8x) & \textbf{49.4} ($+\textbf{3.5}$) & \textbf{31.2} (+\textbf{2.7}) \\
\hline
\multirow{4}{*}{Sem Drone} & Trad  & 83.2 & 51.8 \\
& Trad + ControlAug (80x) & 82.6 ($-0.6$) & 50.1 ($-1.7$) \\
& Trad + InstaGen (80x) & 83.6 ($+0.4$) & 55.3 ($+3.5$) \\
& Trad + Ours (80x) & \textbf{87.4} ($+\textbf{4.2}$) & \textbf{57.7} ($+\textbf{5.9}$) \\
\hline
\end{tabular}
}
\end{center}
\end{table}


\begin{table}[!ht]
\centering
\caption{\textbf{Combining with Existing Augmentations}. mAP on COCO 5k when we apply spatiotemporal augmentations along with existing augmentations. Further performance gains can be obtained by incorporating spatiotemporal augmentations.}
\label{tab:combined-aug}
\resizebox{0.8\columnwidth}{!}{
\begin{tabular}{cccc}
\toprule
 & Raw & ControlAug\cite{fang2024data} & InstaGen\cite{feng2024instagen} \\
\midrule
Base & 24.8 & 27.6 & 28.1 \\
+Ours & 28.6 & 29.7 & 29.1 \\
\midrule
$\Delta$ & +3.8 & + 2.1 & +1.0 \\
\bottomrule
\end{tabular}
}
\end{table}


\begin{table}[t]
\centering
\caption{\textbf{Instance segmentation on COCO.} Our generative augmentation consistently improves instance segmentation performance compared to traditional augmentation alone, yielding gains across different dataset sizes.}
\label{tab:seg}
\resizebox{0.7\columnwidth}{!}{
\begin{tabular}{lcccc}
\toprule
Dataset size & 5k & 10k & 15k & 20k \\
\midrule
Raw & 19.2 & 25.4 & 28.4 & 30.2 \\
Augmented & 22.7 & 27.9 & 30.6 & 32.2 \\
\midrule
$\Delta$ & +3.5 & +2.5 & +2.2 & +2.0 \\
\bottomrule
\end{tabular}
}
\end{table}

\subsection{Effects of Types of Spatiotemporal Augmentation}
The most effective augmentation strategy depends on the dataset's characteristics, as shown in Table~\ref{tab:trajectory}.

For COCO5k, which is object-centric and often static, spatial augmentation with orbital camera trajectories perform the best. This approach exposes novel perspectives that align with the dataset's underlying distribution, improving recall and accuracy.

In contrast, VisDrone scenes are not object-centric, making 3D camera trajectories less suitable for augmentation. 
Many VisDrone images are captured from a bird’s-eye view with the camera pointed steeply downward; 
applying orbital camera motion in such cases can cause the virtual trajectory to intersect the ground plane, producing invalid or unrealistic views.
Here, temporal augmentation with scene dynamics is more effective. By varying object poses and scene states from similar viewpoints, this strategy enriches the dataset with in-distribution temporal variations, boosting recall without sacrificing realism.

These findings show that augmentation must match the dataset's dominant source of variation. Spatial generation directly creates new viewpoints, while temporal generation implicitly provides new information through object motion.


\begin{table}[ht]
    \caption{\textbf{Effects of Spatial vs.\ Temporal augmentation} on COCO5k and VisDrone performance. We use DimensionX \cite{sun2024dimensionx} for spatial generation and LTX-Video-13B \cite{hacohen2024ltx} for temporal generation.}
    \label{tab:trajectory}
    \centering
    \resizebox{\columnwidth}{!}{
    \begin{tabular}{c|c|c|c}
    \hline
    Dataset & Method & mAP@50 & mAP@50:95 \\
    \hline 
    \multirow{2}{*}{COCO5k}
     & Spatial (8x) & \textbf{41.1} (+4.8) & \textbf{28.6} (+3.8)  \\ 
     & Temporal (8x) & 38.7 (+2.4) & 26.7 (+1.9) \\ 
    \hline 
    \multirow{2}{*}{VisDrone}
     & Spatial (40x) & 49.2 (+3.3) & 31.1 (+2.6) \\ 
     & Temporal (40x) & \textbf{49.6} (+3.7) & \textbf{31.3} (+2.8) \\ 
    \hline 
    \end{tabular}
    }
\end{table}

\subsection{Performance Scaling with Dataset Size}

\begin{figure}[t]
    \centering
    \includegraphics[width=\linewidth]{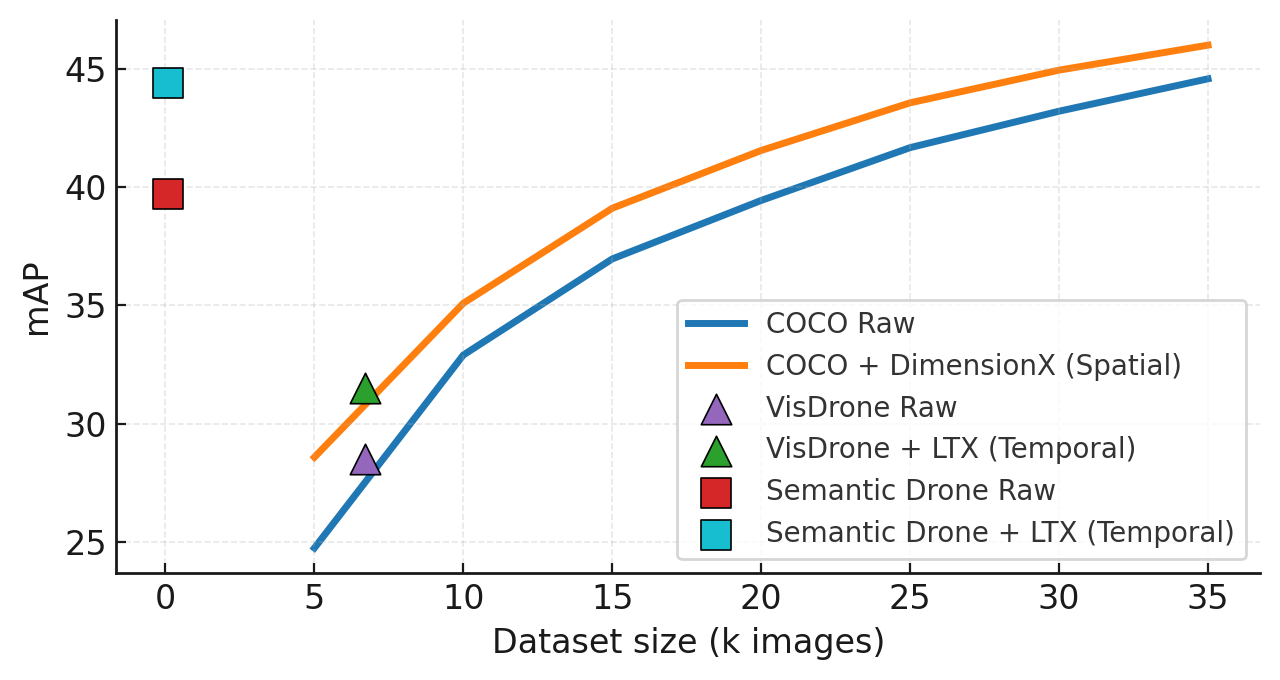}
    \caption{\textbf{Dataset size vs. mAP on 3 datasets.} Detection results of YOLOv11x on COCO under different dataset sizes}
    \label{fig:dataset-size}
\end{figure}

We next study how the benefits of spatiotemporal augmentation scale with dataset size. 
Figure~\ref{fig:dataset-size} plots detection performance as a function of training set size, comparing traditional augmentation and our method. YOLO11x model exhibits a meaningful performance gap that persists even at larger dataset sizes, although the gap decreases with larger sizes.
For VisDrone and Semantic Drone, where the dataset is limited to 6,471 images, we observe impressive  improvements up to $\sim$6\% from our augmentation.  This highlights that in low-data regimes, and particularly in UAV scenarios, spatiotemporal augmentation is especially valuable. Together, these results suggest that our method is most impactful when dataset characteristics allow the model to leverage the expanded coverage it provides. In addition, our augmentation can be applied to detectors with different architectures, which we discuss in the supplementary materials.

\subsection{Ablation Studies}
\label{sec:ablations}
We performed ablation studies on COCO5k to optimize our method's components: frame sampling, data ratios, pseudo-labeling, and data filtering.

\paragraph{Effect of Frame Sampling and Data Ratio.} As shown in Tables~\ref{tab:1_1_sampling} and~\ref{tab:sampling_strategies}, training on raw data yields a 36.3 mAP@50 baseline. We found the optimal strategy is to enforce a 1:1 mixing ratio between real and synthetic data and to sparsely sample 8 frames (e.g. sample every 5 frames). This configuration achieves 41.1 mAP@50 and 28.6 mAP@50:95, a +4.8 gain in mAP@50 over the baseline, and is used as our default.

\paragraph{Effect of Data Filtering.} 
Using this improved 41.1 mAP@50 baseline, we further explore the effect of incorporating a CLIP-based data filtering strategy to remove low-quality generated samples. We follow the data filtering protocol as \cite{fang2024data}. As shown in Table~\ref{tab:data_filtering}, filtering out 10\% of low-quality synthetic samples gives a final slight boost to 41.4 mAP@50 and 29.0 mAP@50:95.

These ablations confirm that 1:1 mixing and sparse frame selection are crucial for achieving optimal performance, while optional data filtering can provide a marginal additional improvement. 




\subsection{Results on Self-supervised learning}
\label{sec:exp-ssl}
We pretrain a ViT-S on COCO35k using DINO for 400 epochs and evaluate on ImageNet 1k via $k$-NN ($k=20$, following the original DINO) and linear probing protocols. As shown in Tab.~\ref{tab:ssl}, our generative spatiotemporal augmentation consistently outperforms standard augmentations across all metrics. Sampling two frames from generated videos yields more diverse, semantically consistent and diverse positive pairs with realistic spatiotemporal variations than relying only on traditional augmentation techniques. Though absolute scores reflect the small pretraining set, the consistent relative gains validate our method's effectiveness for learning transferable representations, especially on the low-data scenarios.

\begin{table}[t]
\centering
\caption{\textbf{Ablation on 1:1 mixing}. Comparison of mAP on COCO5k with and without 1:1 mixing between raw and augmented data. Enforcing balanced data mixing further improves performance.}
\label{tab:1_1_sampling}
\resizebox{\columnwidth}{!}{
\begin{tabular}{lccc}
\toprule
 & raw & w/o 1:1 mixing & w/ 1:1 mixing \\
\midrule
mAP@50 & 36.3 & 39.9 (+3.6) & \textbf{41.1} (+4.8) \\
mAP@50:95 & 24.8 & 27.9 (+3.1) & \textbf{28.6} (+3.8) \\
\bottomrule
\end{tabular}
}
\end{table}

\begin{table}[!ht]
\centering
\caption{\textbf{Effect of different frame selection strategies} on COCO5k performance. Frame selection starts from the first frame of the generated video. The 1:1 mixing ratio is employed for all experiments.}
\label{tab:sampling_strategies}
\resizebox{\columnwidth}{!}{
\begin{tabular}{lccc}
\toprule
Data & \#frames per img \quad & mAP@50 & mAP@50:95 \\
\midrule
Raw & -- & 36.3 & 24.8 \\
Sample every 3 frames & 8 & 40.7 & 28.2 \\ 
Sample every 4 frames & 8 & \textbf{41.1} & 28.5 \\ 
Sample every 5 frames & 8 & \textbf{41.1 }& \textbf{28.6} \\ 
Use first 36 frames & 36 & 40.5 & 28.3 \\ 
\bottomrule
\end{tabular}
}
\end{table}

\begin{table}[!ht]
\centering
\caption{\textbf{CLIP-based filtering.} Performance optimization by filtering augmented data on COCO5k.}
\label{tab:data_filtering}
\resizebox{0.9\columnwidth}{!}{
\begin{tabular}{lccccccc}
\toprule
Removed samples & w/o filtering & 5\% & 10\% & 20\% \\
\midrule
mAP@50 & 41.1 & 40.5 & \textbf{41.4} & 40.4 \\
mAP@50:95 & 28.6 & 28.3 & \textbf{29.0} & 28.2 \\
\bottomrule
\end{tabular}
}
\end{table}

\begin{table}[!ht]
\centering
\caption{\textbf{DINO results}. ImageNet classification performance of a ViT-S pretrained with DINO using generative augmentation. Our augmentation consistently improves both $k$-NN and linear evaluation accuracy.}
\label{tab:ssl}
\setlength{\tabcolsep}{6pt}
\resizebox{0.8\columnwidth}{!}{
\begin{tabular}{lcccc}
\toprule
 & \multicolumn{2}{c}{$k$-NN} & \multicolumn{2}{c}{Linear} \\
Augmentation & Top-1 & Top-5 & Top-1 & Top-5 \\
\midrule
Trad Aug & 23.13 & 38.81 & 27.86 & 49.28 \\
Gen Aug  & 25.87 & 42.20 & 29.90 & 52.39 \\
\midrule
$\Delta$ & +2.74 & +3.39 & +2.04 & +3.11 \\
\bottomrule
\end{tabular}
}
\end{table}

%% file: sec/5_conclusion.tex
\section{Conclusion}
We show that spatiotemporal augmentation with video diffusion models is an effective way to boost perception performance in data-scarce settings. Unlike prior approaches, our method expands data along the critical axes of camera viewpoint and scene dynamics, yielding substantial gains in \textit{low-data regimes} such as UAV-captured imagery. Beyond these improvements, we provide practical guidelines for implementation, including model choice, annotation transfer, and disocclusion handling. Our experiments demonstrate that this approach meaningfully broadens the training distribution and improves performance when annotated data is the bottleneck.

\paragraph{Limitations.} While our method demonstrates strong performance across various settings, several limitations remain. First, the performance gains naturally shrink as the original dataset size grows; therefore, our approach provides maximum utility in data-scarce, high-cost domains (e.g., UAV imagery) rather than in data-abundant regimes. Second, off-the-shelf video diffusion models incur substantial computational costs: generating a single video requires approximately 18 seconds using LTX-Video-13B-distilled and 238 seconds using DimensionX on an NVIDIA L40S GPU. Scaling this approach to massive datasets will require future algorithmic and system-level optimizations. Finally, our automatic annotation pipeline relies on SAM 2, which occasionally drops tracking on very small objects during complex camera motions in dense aerial imagery (e.g., VisDrone). These segmentation errors can introduce false negatives into the generated training set. Addressing these tracking robustness and computational bottlenecks remains an exciting direction for future work.


%% file: sec/X_suppl.tex
\clearpage
\setcounter{page}{1}
\maketitlesupplementary

\section{Implementation Details}

\subsection{Data Generation} 

\paragraph{Spatial Video Generation} For 3D spatial video generation, we use the orbital DimensionX model~\cite{sun2024dimensionx}, which produces videos of a static scene with a camera moving along an orbital trajectory, conditioned on an input image and a text prompt. We obtain the text prompt using a BLIP-2–generated caption~\cite{li2023blip}, and resize and crop the input image to $720 \times 480$, the resolution required by DimensionX.

\paragraph{Temporal Video Generation} For 3D spatial video generation, we use the image-conditioned LTX-Video-13B~\cite{hacohen2024ltx}. We generate an image caption with Florence-2-large \cite{xiao2024florence} and use Llama-3.2-3B~\cite{grattafiori2024llama} to write a video prompt based on the generated caption. 
For COCO~\cite{lin2014microsoft}, each image is first resized using our resolution-adjustment rule: if the image area exceeds $768 \times 576$, we uniformly downscale it so that its area matches this limit, and then round both height and width up to the nearest multiple of 32 to satisfy the model’s resolution constraints. For VisDrone~\cite{cao2021visdrone}, we extract five random $768 \times 576$ crops from each image and generate one 81-frame video clip per crop. From each clip, we sample 8 frames at 5-frame intervals, yielding 40 augmented images per original image.
For Semantic Drone~\cite{Charles2013}, the higher image resolution requires an additional rescaling step: we randomly rescale and crop ten $768 \times 576$ regions per image, generate one video clip for each region, and again sample 8 frames every 5 frames. This produces 80 augmented images per original image.

We use the following prompts to Llama-3.2-3B for video prompt generation on COCO and VisDrone respectively.

\vspace{10pt}
\noindent\textbf{COCO}
\begin{tcolorbox}[breakable, colback=white, colframe=black, arc=0pt]
\textbf{Instruction:}

\noindent Generate a single-paragraph prompt for a video generation model based on a given caption.

\noindent\textbf{Guidelines:}

\begin{itemize}
    \item \textbf{Start with a literal description} of the scene shown in the caption. This visual description is the highest priority.
    \item Follow with \textbf{main action or movement}, then add \textbf{specific gestures, motions, or changes over time}, if present.
    \item Include \textbf{details of gestures, motion}, and \textbf{camera behavior} (e.g., panning, zooming, circling, tracking).
    \item If the caption describes a static scene (e.g., no people or actions), add \textbf{camera movement} (e.g., slow pan, orbit, dolly) to bring the scene to life. Do \textbf{not} simply restate the caption.
    \item \textbf{Do not introduce any new people, objects, or elements} not present in the caption. If the caption has no people, do not describe human actions.
    \item The entire prompt must be \textbf{under 100 words}.
    \item Use \textbf{precise, chronological, and detailed} descriptions.
    \item \textbf{Output only the generated prompt.}
\end{itemize}
\end{tcolorbox}

\noindent\textbf{VisDrone}
\begin{tcolorbox}[breakable, colback=white, colframe=black, arc=0pt]
\textbf{Instruction:}

\noindent Generate a single-paragraph prompt for a video generation model based on a given caption, assuming the video is captured from a \textbf{drone's perspective}.

\noindent\textbf{Guidelines:}

\begin{itemize}
    \item \textbf{Begin with a literal and visual description} of the scene as seen from the drone. Prioritize accurate portrayal of the setting and spatial layout from above or as appropriate for a drone.
    \item Follow with the \textbf{main action or movement}, then describe \textbf{any dynamic elements}, such as motions, gestures, or environmental changes visible from the aerial view.
    \item Include \textbf{details about drone movement}—such as hovering, ascending, descending, panning, circling, tracking, or sweeping—matching the scene's content and emphasizing temporal progression.
    \item If the caption depicts a \textbf{static or natural landscape}, animate the scene using \textbf{drone motion} (e.g., slow orbit, forward glide, downward tilt).
    \item \textbf{Do not add any elements not explicitly present} in the caption (e.g., avoid introducing people, animals, or vehicles not mentioned).
    \item Keep the prompt \textbf{under 100 words}, using \textbf{precise, chronological, and detailed language} suited for a drone-shot video.
    \item \textbf{Output only the generated prompt.}
\end{itemize}
\end{tcolorbox}

\subsection{Disocclusion Handling}

\paragraph{Computing Disocclusion Masks}
For polygon masking, we place a $16 \times 16$ grid of points on the first frame and track them through the video using TAPNext~\cite{zholus2025tapnext}. For each generated frame, we compute the convex hull of the tracked points and mask out all pixels outside this polygon.

For dense masking, we use AllTracker~\cite{harley2025alltracker} to track every pixel in the first frame throughout the video. For each generated frame, we compute a density map by counting tracked points in each pixel’s neighborhood, weighted by a Gaussian kernel. We obtain the disocclusion mask by thresholding this map, masking pixels whose density falls below a specified threshold, indicating newly exposed or generated regions. See Alg.~\ref{alg:mask} for details.

\begin{algorithm}[t]
\caption{Dense Disocclusion Mask Computation}
\label{alg:mask}
\begin{algorithmic}[1]

\Require Video frames $V = \{I_1, \ldots, I_T\}$; visibility threshold $\tau_{\text{vis}}$; confidence threshold $\tau_{\text{conf}}$; Gaussian std $\sigma$; weight threshold $\tau_w$

\State Convert frames to tensor of shape $(T, H, W, 3)$
\State Generate pixel grid $G \in \mathbb{R}^{1 \times H \times W \times 2}$

\State Run flow model to obtain:
\[
\text{flows},\; \text{visconf} \gets \text{Model.forward\_sliding}(V)
\]

\State Compute pixel trajectories:
\[
\text{traj\_maps} = \text{flows} + G
\]


\State Compute
\[
\text{visibs} \in \{0,1\}^{H \times W \times T}
\]
using thresholds $\tau_{\text{vis}}$ and $\tau_{\text{conf}}$

\For{$t = 1$ to $T$}

    \State Select visible points:
    \[
    P_t = \{(x,y) \mid \text{visibs}[t,i,j] = 1\}
    \] \qquad where \((x,y)=\text{traj\_maps}[t,i,j]\)

    \State Initialize impulse image $M_t \in \mathbb{R}^{H \times W}$ to zeros
    \State For each $(x,y) \in P_t$, set $M_t[y,x] \gets 1$

    \State Construct Gaussian kernel $K$ with radius $3\sigma$

    \State Compute density map:
    \[
    W_t = \mathrm{Conv2D}(M_t, K)
    \]

    \State Threshold to obtain disocclusion mask:
    \[
    \text{mask}_t = (W_t \ge \tau_w)
    \]


\EndFor

\State \Return $\{\text{mask}_t\}_{t=1}^T$

\end{algorithmic}
\end{algorithm}

\paragraph{Pseudo-labeling} For pseudo-labeling, we apply a simple two-pass pseudo-labeling scheme. We first train a detector on the augmented dataset without masking, then run inference on the training set and retain predictions that satisfy all of the following criteria:
\begin{itemize}
\item Confidence exceeds a specified threshold.
\item The proportion of masked area (from the dense disocclusion mask) within the predicted bounding box exceeds a given ratio of the box area.
\item IoU with any ground truth box (from SAM 2) is below a threshold, to avoid duplicate labels.
\end{itemize}

In our experiments, we set both the masked-area ratio threshold and the IoU threshold to 0.5.

\subsection{Training Settings}

We utilize the YOLO11 models implemented in the Ultralytics~\cite{yolo11_ultralytics} framework. The models are trained from scratch on the target dataset. All experiments are conducted with an input resolution of $640 \times 640$ pixels and a batch size of 64. We employ Stochastic Gradient Descent (SGD) as the optimizer with a momentum of 0.937 and a weight decay of $5 \times 10^{-4}$. The initial learning rate is set to $\text{lr}_0 = 0.01$ with a linear decay schedule to a final learning rate factor of $\text{lrf} = 0.01$. We train the models for 400 epochs (800 epochs for YOLO11n to ensure convergence) with a warm-up period of 3 epochs. We apply YOLO11's standard data augmentation pipeline to all experiments, including Mosaic augmentation (probability 1.0), horizontal flipping (probability 0.5), translation ($\pm 10\%$), scaling ($\pm 50\%$), and color space augmentations in the HSV domain with hue (0.015), saturation (0.7), and value (0.4) fractions. 





\section{Tranining with RT-DETR}
We also apply our method to the Transformer-based end-to-end detector, RT-DETR~\cite{lv2023detrs}, to demonstrate that our approach generalizes well across different architectures.

We utilize the RT-DETR-x model implemented in the Ultralytics~\cite{yolo11_ultralytics} framework. The model is trained from scratch without pretraining. All experiments are conducted with an input resolution of $640 \times 640$ pixels and a batch size of 64. 

\begin{figure}[htbp]
    \centering
    \includegraphics[width=\linewidth]{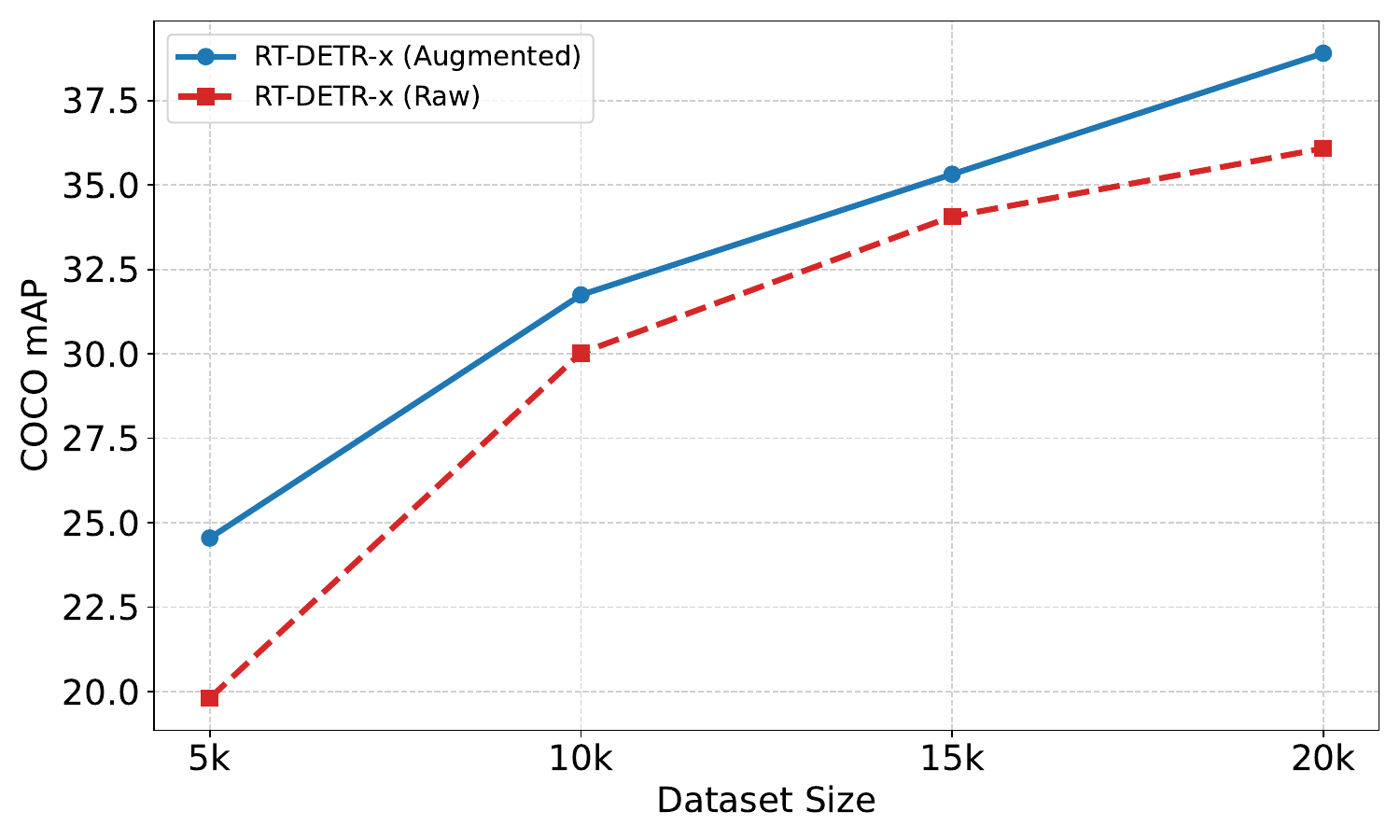}
    \caption{Detection results of RT-DETR-x on COCO under different dataset sizes}
    
    \label{fig:detr}
\end{figure}

As shown in Figure~\ref{fig:detr} and Table~\ref{tab:detr_table}, training with our method consistently outperforms the baseline. This confirms that our method can be robustly applied to diverse model architectures, including both CNN-based and Transformer-based models. Notably, we observe that when data is scarce (5k samples), our method outperforms the baseline by a significant margin. We hypothesize that this is due to the data-hungry nature of Transformers; therefore, generating additional data in data-scarce regimes yields a substantial boost in performance.

\begin{table}[ht]
\centering
\caption{Comparison of mAP@50:95 for \textbf{RT-DETR-x} trained on raw vs. augmented data as training set size increases.}
\label{tab:detr_table}
\resizebox{\columnwidth}{!}{
\begin{tabular}{ccccc}
\toprule
Dataset size & 5k & 10k & 15k & 20k \\
\midrule
Raw & 19.8 & 30.0 & 34.1 & 36.1 \\
Augmented & 24.5 (+4.7) & 31.7 (+1.7) & 35.3 (+1.2) & 38.9 (+2.8) \\
\bottomrule
\end{tabular}
}
\end{table}

\section{Additional Analyses}

\subsection{Effects of Model Capacity}

\begin{figure}[htbp]
    \centering
    \includegraphics[width=\linewidth]{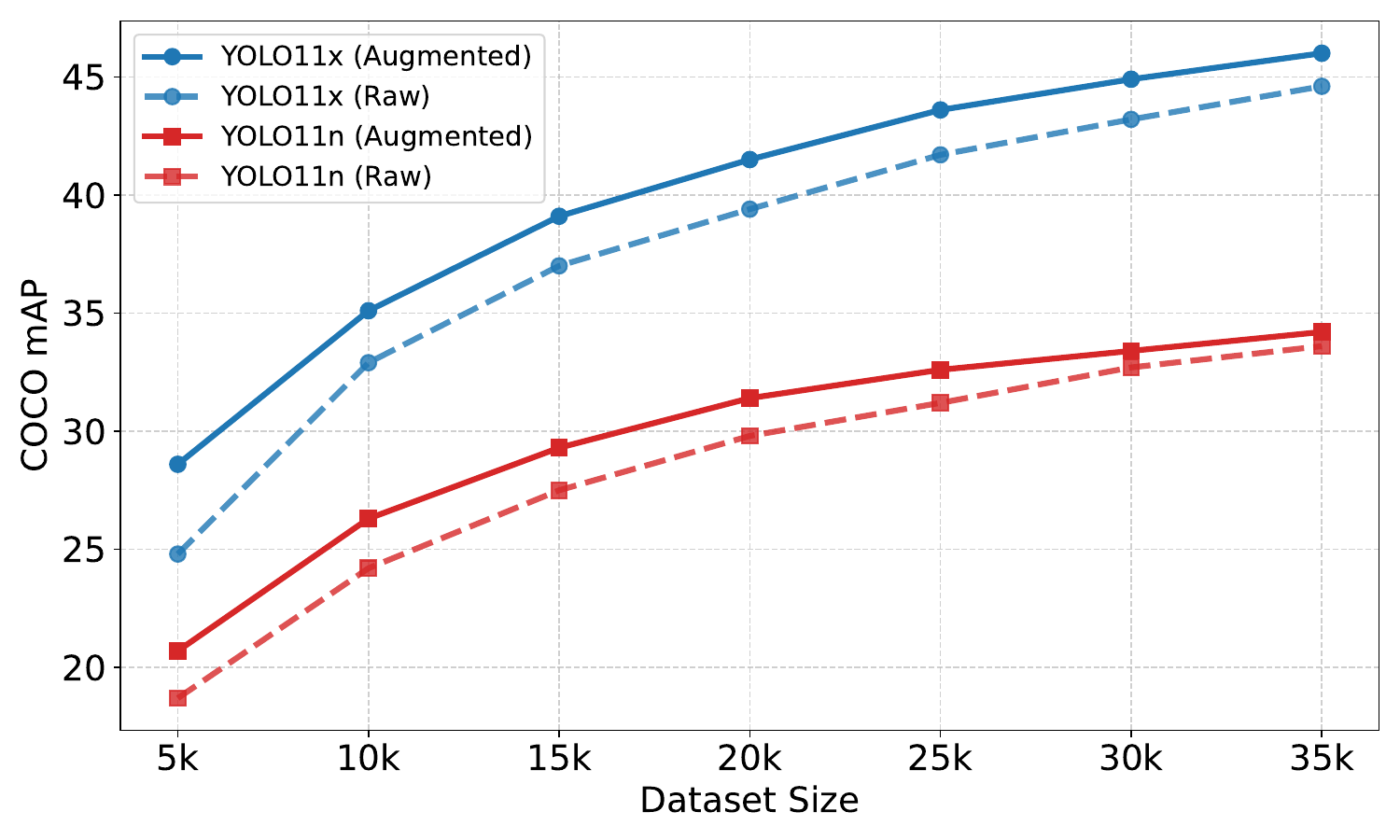}
    
    \caption{\textbf{Performance vs Model Capacity.} Detection results of YOLO11x and YOLO11n on COCO under different dataset sizes}
    
    \label{fig:yolo11n}
\end{figure}

To investigate how the effectiveness of our augmentation method scales with model capacity, we also train a smaller YOLO11n~\cite{yolo11_ultralytics} on raw and augmented COCO datasets under different dataset sizes. As shown in Fig.~\ref{fig:yolo11n}, the effect depends strongly on model capacity. For the low-capacity YOLO model, the initial gap between the two augmentation strategies is small and diminishes further as training data increases, suggesting that limited-capacity models quickly saturate regardless of augmentation. In contrast, the high-capacity YOLO model exhibits a much larger performance gap that persists even at larger dataset sizes, indicating that stronger models can exploit the additional coverage provided by spatiotemporal augmentation.  

\begin{table}[ht]
\centering
\caption{Comparison of mAP@50:95 for \textbf{YOLO11n} trained on raw vs. augmented data as training set size increases.}
\label{tab:yolo11n}
\resizebox{\columnwidth}{!}{
\begin{tabular}{cccccccc}
\toprule
Dataset size & 5k & 10k & 15k & 20k & 25k & 30k & 35k \\
\midrule
Raw & 18.7 & 24.2 & 27.5 & 29.8 & 31.2 & 32.7 & 33.6 \\
Augmented & 20.7 & 26.3 & 29.3 & 31.4 & 32.6 & 33.4 & 34.2 \\
\bottomrule
\end{tabular}
}
\end{table}

\subsection{Experiments on Disocclusion Handling}
\label{sec:exp-disocclusion}
Viewpoint changes introduce disocclusions, where newly revealed areas that may contain unlabeled objects. We explore two approaches to mitigate the resulting supervision gap: masking newly generated regions and generating pseudo-labels for objects appearing in those regions.

\paragraph{Masking disoccluded regions.} We compare three strategies for handling these regions: (i) \emph{w/o masking} (use frames as-is), (ii) \emph{polygon masking} that preserves the tracked, original field-of-view while discarding newly revealed areas, and (iii) \emph{pixel-wise masking} (dense, pixel-level masking of all disoccluded pixels). 
As shown in Table~\ref{tab:masking}, neither of the two masking strategies improves detection performance, and both even reduce accuracy on COCO5k. 

We find that this degradation occurs partly because viewpoint changes in some generated frames reveal previously occluded portions of labeled objects. These objects will be partially masked, making it difficult for the model to learn accurate localization. Meanwhile, the masks produced by SAM 2 naturally extend into newly visible (disoccluded) regions, accurately capturing the full boundary of the object in these frames. Because masking out such regions would discard useful visual information, we omit disocclusion masking from our final augmentation pipeline.

\paragraph{Pseudo-labeling.} We apply a simple two-pass pseudo-labeling scheme. We first train a detector on the augmented dataset without masking, then run inference on the training set and retain predictions above a specified confidence threshold that fall within the newly generated regions. These predictions serve as pseudo-labels for a second round of training. As shown in Table~\ref{tab:conf_threshold}, a confidence threshold of 0.7 yields a modest improvement of 0.3 mAP@50:95.

\begin{table}[t]
\centering
\caption{\textbf{Effect of confidence threshold on pseudo-label} to performance on COCO5k. }
\label{tab:conf_threshold}
\resizebox{\columnwidth}{!}{
\begin{tabular}{lccccc}
\toprule
Conf thr. & 0.5 & 0.6 & 0.7 & 0.8 & w/o pseudo-labels \\
\midrule
mAP@50 & 41.1 & 40.7 & \textbf{41.4} & 40.7 & 41.1 \\
mAP@50:95 & 28.7 & 28.3 & \textbf{28.9} & 28.3 & 28.6 \\
\bottomrule
\end{tabular}
}
\end{table}

\begin{table}[h!]
\caption{\textbf{Ablation on disocclusion masking.} We compare object detection performance with and without polygon and pixel-wise masking on COCO5k and VisDrone datasets.}
\label{tab:masking}
\begin{center}
\resizebox{\columnwidth}{!}{
\begin{tabular}{c|c|c|c}
\hline 
Dataset & Method & mAP@50 & mAP@50:95 \\
\hline 
\multirow{3}{*}{COCO5k} & w/o masking  & \textbf{41.1} & \textbf{28.6} \\
 &  polygon masking & 40.4 & 28.0 \\
  & pixel-wise masking & 39.6 & 27.4 \\
\hline 
\multirow{3}{*}{VisDrone} & w/o masking  & \textbf{49.6} & \textbf{31.3 } \\
 &  polygon masking & \textbf{49.6} & \textbf{31.3}\\
  & pixel-wise masking & 49.5  & 31.1 \\
\hline 
\end{tabular}
}
\end{center}
\end{table}

\paragraph{Practical Recommendation.}
Based on the results in Tab.~\ref{tab:masking},\ref{tab:conf_threshold}, both masking and pseudo-labeling yield only marginal improvements (typically under one mAP point). 
Unless maximizing every fraction of performance, disocclusion can be safely ignored: its impact is minor compared to the overall benefits of spatiotemporal augmentation, and handling it adds extra complexity with limited return.

\subsection{Performance vs Dataset Size on Semantic Drone}

To investigate the effectiveness of our method on Semantic Drone \cite{Charles2013} when dataset size changes, we additionally train on 100 and 300 images, and evaluate on the remaining 100 images. As shown in Tab.~\ref{tab:semantic}, the benefit of our augmentation diminishes at this scale. Since the task is relatively simple, the baseline model is already near saturation. Nonetheless, our spatiotemporal augmentation still yields a 2.1-point improvement in mAP@50:95.

\begin{table}[ht]
\centering
\caption{Comparison of mAP on subsets of \textbf{Semantic Drone}~\cite{Charles2013}. We evaluate YOLO11x detectors trained with and without augmented data using our method on subsets size of 100, 200, and 300, respectively. All models are evaluated on the remaining 100 images.}
\label{tab:semantic}
\resizebox{\columnwidth}{!}{
\begin{tabular}{ccccc}
\toprule
Training set size & Augmented & mAP@50 & mAP@50:95 \\
\midrule
100 &  & 71.1 & 39.4 \\
100 & \checkmark & 74.5 (+3.4) & 44.1 (+4.7) \\
200 &  & 81.9 & 51.3 \\
200 & \checkmark & 86.8 (+4.9) & 56.8 (+5.5) \\
300 &  & 89.8 & 61.3 \\
300 & \checkmark & 90.7 (+0.9) & 63.4 (+2.1) \\
\bottomrule
\end{tabular}
}
\end{table}

\subsection{Effects on Pretrained Models}

Pretraining on large, diverse datasets such as COCO is a standard strategy for improving detection performance in data-scarce domains like drone datasets. Our spatiotemporal augmentation method remains effective even when applied in this pretraining–finetuning pipeline. As shown in Table~\ref{tab:pretrained}, finetuning a COCO-pretrained YOLO11x model on VisDrone with our augmented data yields consistent gains over both the raw-data baseline and the pretrained baseline, improving mAP@50 by 2.0 points and mAP@50:95 by 1.6 points. Although the improvements are slightly smaller than those observed when training from scratch, which reflects the strong initialization provided by COCO pretraining, the results demonstrate that our approach complements pretrained representations rather than duplicating them, providing meaningful additional signal during finetuning.

\begin{table}[ht]
\centering
\caption{\textbf{Effectiveness on Pretrained Models}. We finetune a COCO-pretrained YOLO11x model on VisDrone with and without our augmentation method. Our model can still improve performance significantly in the finetuning setting. Traditional augmentations are applied to all experiments.}
\label{tab:pretrained}
\resizebox{\columnwidth}{!}{
\begin{tabular}{lccc}
\toprule
Data & Pretrained & mAP@50 & mAP@50:95 \\
\midrule
Raw & & 45.9 & 28.5 \\
Augmented (40x) & & 49.6 (+3.7) & 31.3 (+2.8) \\
Raw & \checkmark & 48.2 & 30.1 \\
Augmented (40x) & \checkmark & 50.2 (+2.0) & 31.7 (+1.6) \\
\bottomrule
\end{tabular}
}
\end{table}

\subsection{Ablation on Random Crop on VisDrone}

Due to the large scene scale and small object size in aerial imagery, directly conditioning the video diffusion model on full-resolution images often leads to suboptimal synthesis quality. To address this, we provide the video diffusion model with randomly cropped regions of each image, allowing it to focus on local content. We conduct an ablation study to verify that such random cropping improves downstream detection performance.

\begin{table}[ht]
\centering
\caption{\textbf{Ablation on random cropping for VisDrone.} We train YOLO11x on augmented VisDrone datasets using frames from videos generated by LTXV-13B from resized full images or from randomly cropped regions, respectively. The table reports mAP@50 and mAP@50:95, showing that random cropping consistently improves detection performance, with larger numbers of crops yielding the strongest gains.}
\label{tab:randcrop}
\resizebox{\columnwidth}{!}{
\begin{tabular}{lcc}
\toprule
Data & mAP@50 & mAP@50:95 \\
\midrule
Raw & 45.9 & 28.5 \\
Resize (8x) & 48.2 (+2.3) & 30.3 (+1.8) \\
Random crop (8x) & 48.9 (+3.0) & 30.7 (+2.2) \\
Random crop (40x) & 49.6 (+3.7) & 31.3 (+2.8) \\
\bottomrule
\end{tabular}
}
\end{table}

As shown in Tab.~\ref{tab:randcrop}, using frames generated from resized full images yields a clear improvement over training on raw data alone (+2.3 mAP@50 and +1.8 mAP@50:95). However, replacing the resized inputs with randomly cropped regions leads to consistently stronger gains under the same number of augmentation images (+3.0 and +2.2), confirming that localized conditioning produces higher-quality and more informative synthetic views. Increasing the number of random crops further amplifies this effect: using 40x augmentation (5 crops per original image) pushes performance to 49.6 mAP@50 and 31.3 mAP@50:95, corresponding to improvements of +3.7 and +2.8 over the baseline. These results demonstrate that random cropping not only enhances generative quality but also translates into tangible detector gains.

\section{Computation and Performance Comparison}

Table~\ref{tab:compute} compares the total augmentation time per source image with downstream detection performance on VisDrone. InstaGen, ControlAug, our method with LTXV-13B-distilled, and our method with DimensionX require 45, 169, 110, and 1,210 seconds per source image and achieve 29.8, 30.5, 31.3, and 31.1 mAP, respectively. The reported times for our method include both video generation and SAM~2 label propagation. Although video diffusion is generally more expensive than image-based augmentation, LTXV-13B-distilled provides a competitive balance between computational cost and quality, outperforming the other methods while remaining substantially faster than DimensionX. We observed that simply using a smaller video model often reduces temporal consistency. For example, object motion or identity may become inconsistent across frames, making the generated videos unsuitable for reliable object tracking. Therefore, when the computation budget is limited, we recommend inference-optimized video diffusion models rather than arbitrarily smaller generators. In particular, LTXV-13B-distilled uses only eight diffusion sampling steps while producing generation quality comparable to the full LTXV-13B model.

\begin{table}[ht]
  \centering
  \footnotesize
  \setlength{\tabcolsep}{4pt}
  \renewcommand{\arraystretch}{0.9}
  \begin{tabular}{lcccc}
    \toprule
    Method             & ControlAug & InstaGen       & \textbf{Ours-LTX} & \textbf{Ours-DimX} \\
    \midrule
    mAP                & 30.5       & 29.8           & \textbf{31.3}     & 31.1     \\
    $\Delta$           & +2.0       & +1.3           & \textbf{+2.8}     & +2.6     \\
    Gen\ time/src-img & 169\,s     & \textbf{45\,s} & 110\,s            & 1210\,s  \\
    \bottomrule
  \end{tabular}
  \vspace{-0.2cm}
  \caption{VisDrone mAP vs.\ wall-clock cost per source image. Our method attains the largest improvement at a competitive cost. For our entries, the reported time covers generating 5 videos per source image and includes SAM-2 mask propagation (4\,s per video).}
  \vspace{-0.1cm}
  \label{tab:compute}
\end{table}